\documentclass{ieeeaccess}
\usepackage{cite}
\usepackage{amsmath,amssymb,amsfonts}
\usepackage{algorithmic}
\usepackage{graphicx}
\usepackage{textcomp}
\def\BibTeX{{\rm B\kern-.05em{\sc i\kern-.025em b}\kern-.08em
    T\kern-.1667em\lower.7ex\hbox{E}\kern-.125emX}}

\usepackage{hyperref} 
\hypersetup{
    colorlinks=true,
    linkcolor=blue,
    citecolor=blue,     
    urlcolor=cyan,
    pdfpagemode=FullScreen,
    }
\usepackage{tabularx}
\newcolumntype{Y}{>{\centering\arraybackslash}X}  
\newcolumntype{R}{>{\raggedleft\arraybackslash}X} 
\newcolumntype{L}{>{\raggedright\arraybackslash}X} 
\usepackage{booktabs}

\usepackage{xcolor}

\makeatletter
\colorlet{accessblue}{black}
\def\@maketitle{%
     \setbox\maketitlebox\hbox\bgroup%
     \begin{minipage}{\titlewidth}%
     \raggedright\vspace*{1.2mm}%
     \ifaccesseditors\else{\vss\historyfont\@history\vss}\vspace*{2mm}\par\fi%
     {\vss\color{accessblue}\titlefont\@title\vss}\vspace*{6mm}\par%
     \ifaccesseditors\vspace*{2mm}\else{\vss\authorfont\@author\vss}\vspace*{0mm}\par%
     \@tempcnta=0%
     \loop\ifnum\@tempcnta<\theaddrCtr%
       \advance\@tempcnta by 1%
       \par{\vss\addressfont\textsuperscript{\expandafter\csname addrInd\the\@tempcnta\endcsname}
        \expandafter\csname addr\the\@tempcnta\endcsname\vss}%
     \repeat\vspace*{1mm}\par%
     {\vss\correspfont\@corresp\vss\vspace{2mm}}\par\fi%
     {\vss\tfootnotefont\@tfootnote\vss\vspace{7mm}}\par%
     \noindent\box99\vspace{6mm}\par%
     \ifaccesseditors\vspace*{8mm}\else%
     \noindent\box88\vspace{18mm}\par%
     \fi%
     \end{minipage}\par%
     \egroup%
     \fixSkip{\the\ht\maketitlebox}%
}

\def\ps@titlepage{
  \def\@oddhead{} 
  \def\@evenhead{} 
  \def\@oddfoot{\reset@font\hfil\thepage\hfil} 
  \def\@evenfoot{\reset@font\hfil\thepage\hfil} 
}

\def\ps@headings{
  \def\@oddhead{\parbox[t]{\textwidth}{\mbox{}\\[-6.3mm]{\raisebox{-2pt}{\headerfont\rightmark}}\hfill\mbox{}\vspace*{-1.5mm}\par\hrulefill}}
  \def\@evenhead{\parbox[t]{\textwidth}{\mbox{}\\[-6.3mm]\mbox{}\hfill{\raisebox{-2pt}{\headerfont\leftmark}}\vspace*{-1.5mm}\par\hrulefill}}
  
  \def\@oddfoot{\reset@font\hfil\thepage\hfil}
  \def\@evenfoot{\reset@font\hfil\thepage\hfil}
}

\makeatother

\begin{document}

\title{Masked Autoencoders for Ultrasound Signals: Robust Representation Learning for Downstream Applications\\
}
\author{\uppercase{Immanuel Roßteutscher}\authorrefmark{1},
\uppercase{Klaus S. Drese\authorrefmark{1},
and Thorsten Uphues}\authorrefmark{1}}
\address[1]{
Institute for Sensor and Actuator Technology, Coburg University of Applied Sciences and Arts, Am Hofbräuhaus 1B, 96450 Coburg, Germany (e-mail: immanuel.rossteutscher@hs-coburg.de)}

\markboth
{Immanuel Roßteutscher et al.: \headeretal: Masked Autoencoders for Ultrasound Signals}
{Immanuel Roßteutscher et al.: \headeretal: Masked Autoencoders for Ultrasound Signals}

\corresp{Corresponding author: Immanuel Roßteutscher (e-mail: immanuel.rossteutscher@hs-coburg.de).}

\begin{abstract}
We investigated the adaptation and performance of Masked Autoencoders (MAEs) with Vision Transformer (ViT) architectures for self-supervised representation learning on one-dimensional (1D) ultrasound signals. Although MAEs have demonstrated significant success in computer vision and other domains, their use for 1D signal analysis, especially for raw ultrasound data, remains largely unexplored. Ultrasound signals are vital in industrial applications such as non-destructive testing (NDT) and structural health monitoring (SHM), where labeled data are often scarce and signal processing is highly task-specific. We propose an approach that leverages MAE to pre-train on unlabeled synthetic ultrasound signals, enabling the model to learn robust representations that enhance performance in downstream tasks, such as time-of-flight (ToF) classification. This study systematically investigated the impact of model size, patch size, and masking ratio on pre-training efficiency and downstream accuracy. Our results show that pre-trained models significantly outperform models trained from scratch and strong convolutional neural network (CNN) baselines optimized for the downstream task. Additionally, pre-training on synthetic data demonstrates superior transferability to real-world measured signals compared with training solely on limited real datasets. This study underscores the potential of MAEs for advancing ultrasound signal analysis through scalable, self-supervised learning.
\end{abstract}

\titlepgskip=-15pt
\maketitle

\section{Introduction}\label{sec:introduction}
Transformer models have revolutionized the fields of Natural Language Processing (NLP) and Computer Vision (CV), achieving significant progress in different tasks \cite{vaswani_attention_2017, dosovitskiy_image_2021}. A key factor in this success is the capability of transformers to perform exceptionally well in self-supervised pre-training, which enables models to learn meaningful representations from large amounts of unlabeled data. Transformer-based architectures, such as BERT \cite{devlin_bert_2019} and GPT \cite{radford_improving_2018} in NLP and BEiT \cite{bao_beit_2022} in CV have demonstrated impressive generalization capabilities across various downstream tasks, achieving high performance without extensive task-specific supervision. This self-supervised approach has also been successfully applied to a wide range of other domains, including bioinformatics \cite{ji_dnabert_2021}, speech processing \cite{baevski_wav2vec_2020}, finance \cite{yang_finbert_2020}, and astronomy \cite{morvan_dont_2022}, demonstrating its versatility.

However, adapting self-supervised representation learning with transformer architectures to one-dimensional (1D) ultrasound signals remains largely unexplored despite its potential for practical applications. Ultrasound-based sensors are widely used in various industrial applications, including nondestructive testing, structural health monitoring \cite{hassani_systematic_2023}, \cite{brunner_structural_2021}, \cite{kot_recent_2021} and time-of-flight (ToF) estimation tasks such as distance measurement, level detection, and flow monitoring \cite{qiu_review_2022}, \cite{gao_analysis_2021}, \cite{ren_design_2022}, \cite{jagatheesaperumal_comprehensive_2020}. For each specific application, a separate signal processing procedure typically needs to be developed, which is both time- and resource-intensive. Furthermore, labeling large amounts of ultrasound data is costly and often requires expert knowledge, posing a significant barrier to the adoption of deep-learning-based solutions. A more generalizable approach, based on learned representations from unlabeled raw ultrasound signals, could significantly reduce the need for custom-designed solutions and improve the efficiency of downstream tasks in this domain. To address these challenges, self-supervised Masked Image Modeling (MIM) approaches \cite{bao_beit_2022}, \cite{he_masked_2022}, \cite{xie_simmim_2022} particularly Masked Autoencoders (MAEs) \cite{he_masked_2022}, offer promising solutions. The following section examines why this approach aligns well with the unique properties of ultrasound signals and their industrial applications.

While both ultrasound signals and language share a one-dimensional structure, the characteristics of ultrasound signals align more closely with those of images, making CV approaches more applicable. In contrast to the discrete nature of words \cite{xie_simmim_2022}, images and ultrasound signals are continuous data without a fixed vocabulary. Additionally, language represents a high-level, human-designed concept with a high information density \cite{he_masked_2022}, whereas both images and ultrasound signals naturally capture physical phenomena without predefined structure.

Given these characteristics, self-supervised MIM approaches are particularly relevant to ultrasound data. Among these, MAE architecture is a highly effective approach. During pre-training, unlike other MIM methods, only the unmasked patches are fed into the encoder of the asymmetric encoder-decoder structure \cite{he_masked_2022}. Combined with a high masking ratio that has proven advantageous for learning meaningful representations, especially in images, this approach significantly reduces computational costs, which is a key advantage for industrial applications. Moreover, the comparatively simple model architecture, which does not rely on discrete visual tokens \cite{bao_beit_2022}, reduces complexity, thereby lowering the implementation effort in industrial settings and potentially increasing acceptance. Finally, the use of a standard Vision Transformer (ViT) backbone \cite{dosovitskiy_image_2021}, which processes vector representations of flattened image patches, enables straightforward adaptation to ultrasound signals without modifying the overall architecture.

The objective of this study was to investigate the potential of MAE with a ViT backbone for self-supervised representation learning on one-dimensional ultrasound signals. By leveraging pre-training on unlabeled synthetic data, this approach aims to address key challenges in industrial ultrasound applications, such as the limited availability of labeled data and reliance on task-specific signal processing.

To achieve this, the study systematically evaluated key hyperparameters, including model size, patch size, and masking ratio, to optimize performance during pre-training and downstream tasks. Furthermore, it demonstrates the generalization capabilities of models pre-trained on synthetic data, achieving significant improvements when applied to real-world ultrasound signals compared to models trained from scratch or pre-trained on measured data, and strong Convolutional Neural Network (CNN) baselines optimized for the task.

\section{Related Work}
\subsection{MAE for Image Analysis}
Self-supervised pre-training methods for transformers in computer vision have evolved from early approaches, such as iGPT \cite{chen_generative_2020}, which used autoregressive modeling of image pixels, and DINO \cite{caron_emerging_2021}, which relied on self-distillation, to more recent developments that focus on MIM. MIM extends the principles of Masked Language Modeling (MLM) to images by masking randomly selected image patches and training the model to reconstruct the missing content based on the information available from the unmasked patches. Leading methods in MIM include BEiT \cite{bao_beit_2022}, which predicts discrete visual tokens obtained from a pre-trained tokenizer, as well as MAE \cite{he_masked_2022} and SimMIM \cite{xie_simmim_2022}, both of which directly reconstruct pixel values of masked patches.

Although MAE and MAE-like architectures remain active areas of research, They have already demonstrated their strengths across a variety of domains and are regarded as one of the most promising approaches in MIM \cite{zhou_masked_2023}. Beyond its applications in real-world image processing \cite{liu_pose_2022}, \cite{ma_advanced_2022}, \cite{xu_transfer_2022}, MAE architecture has achieved significant advancements in more specialized fields. These include medical image analysis for tasks such as classification and segmentation \cite{wang_cross-attention_2023}, \cite{qin_self-supervised_2022}, \cite{huang_self-supervised_2023}, \cite{huang_efficient_2022}, \cite{xiao_delving_2023}, \cite{cao_unsupervised_2024}, \cite{zhou_self_2023}, geographic applications \cite{sosa_how_2024}, \cite{wang_land_2022}, \cite{fuller_satvit_2022}, and spatiotemporal masking and feature learning, as demonstrated in video anomaly detection \cite{hu_detecting_2022, datchanamoorthy_anomaly_2022} and action classification \cite{feichtenhofer_masked_2022}.

\subsection{MAE for One-Dimensional Data}
In contrast to images, MAE research lags behind in the area of one-dimensional data such as time series. Nevertheless, several studies have applied MAEs to time-series forecasting. For example, MTSMAE \cite{tang_mtsmae_2022} introduced a self-supervised pre-training approach for time-series data, demonstrating improved performance over traditional supervised methods across various datasets. These include datasets such as ETT, which records the oil temperature and power load, and WTH, which provides hourly weather data. Another example is provided in the paper TI-MAE: Self-Supervised Masked Time Series Autoencoders \cite{li_ti-mae_2023}, which addresses challenges in time-series forecasting by introducing an MAE framework. Unlike contrastive learning approaches, TI-MAE directly reconstructs masked embedded time series inputs at the point level. The framework demonstrated an improved performance in forecasting and classification tasks across several real-world datasets. Further examples can be found in the field of electroencephalography (EEG representation learning), where MAE has been adapted to address challenges such as limited labeled datasets. GMAEEG \cite{fu_gmaeeg_2024} leverages a graph-based MAE to reconstruct spatial and temporal EEG features, excelling in tasks such as emotion and disease recognition. Similarly, MAEEG \cite{chien_maeeg_2022} applied an MAE framework to enhance sleep stage classification and achieved significant gains with limited labeled data.

\subsection{MAE for Ultrasound Signals}
To the best of our knowledge, no studies have explored the application of MAE to one-dimensional ultrasound raw signals. The closest work, Wavefield MAE \cite{ye_wavefield_2024}, focused on analyzing two-dimensional (2D) ultrasonic wavefield data using a two-stage adaptation of large vision models, demonstrating the potential of MAE in reconstructing and classifying wavefield patterns. However, this study targets spatially resolved 2D data, which differs significantly from the 1D nature of raw ultrasound signals.

\newpage

\section{Methodology}\label{sec:methodology}
\subsection{MAE Architecture}
As shown in Fig.~\ref{fig:vitmae}, we used the MAE architecture \cite{he_masked_2022} with a large encoder, which processes only unmasked 1D signal patches to learn representations of the ultrasound signals. The smaller decoder, dedicated solely to signal reconstruction, utilizes information from the contextualized visible patches to predict the masked regions based on the latent representations. After pre-training, the fine-tuning model retains only the encoder, leveraging the learned features of the ultrasound signals for downstream tasks.

Let the input signal be $x \in \mathbb{R}^{L}$, divided into $N$ non-overlapping patches $x_p = \{x_{p,1}, \dots, x_{p,N}\}$ where $x_{p,i} \in \mathbb{R}^{P}$ and $L=N \cdot P$. During pre-training, a subset of patches with indices $\mathcal{M} \subset \{1, \dots, N\}$ is masked. The MAE encoder $E$ processes only the visible patches $\{x_{p,i} \mid i \notin \mathcal{M}\}$ along with their positional embeddings, yielding latent representations $z = \{z_{p,i} \mid i \notin \mathcal{M}\}$. The decoder $D$ receives a full sequence constructed from these latent representations and specific mask tokens ($m_{t,j}$ for $j \in \mathcal{M}$ as shown in Fig.~\ref{fig:vitmae}), both of which incorporate positional information. Based on this input sequence, the decoder generates a full sequence of reconstructed patches $\hat{x}_p = \{\hat{x}_{p,i} \mid i \in \{1, \dots, N\}\}$. The objective is to minimize the reconstruction error, which, in our case, is the Mean Absolute Error (MAE) computed only for the originally masked patches:
\begin{equation} \label{eq:mae_loss}
\mathcal{L}_{\text{MAE}} = \frac{1}{|\mathcal{M}| \cdot P} \sum_{i \in \mathcal{M}} \| x_{p,i} - \hat{x}_{p,i} \|_1 
\end{equation}
where $\| \cdot \|_1$ denotes the L1 norm, summing the absolute differences element-wise across all samples in the masked patches.

\Figure[ht!](topskip=0pt, botskip=0pt, midskip=0pt)[width=0.999\columnwidth]{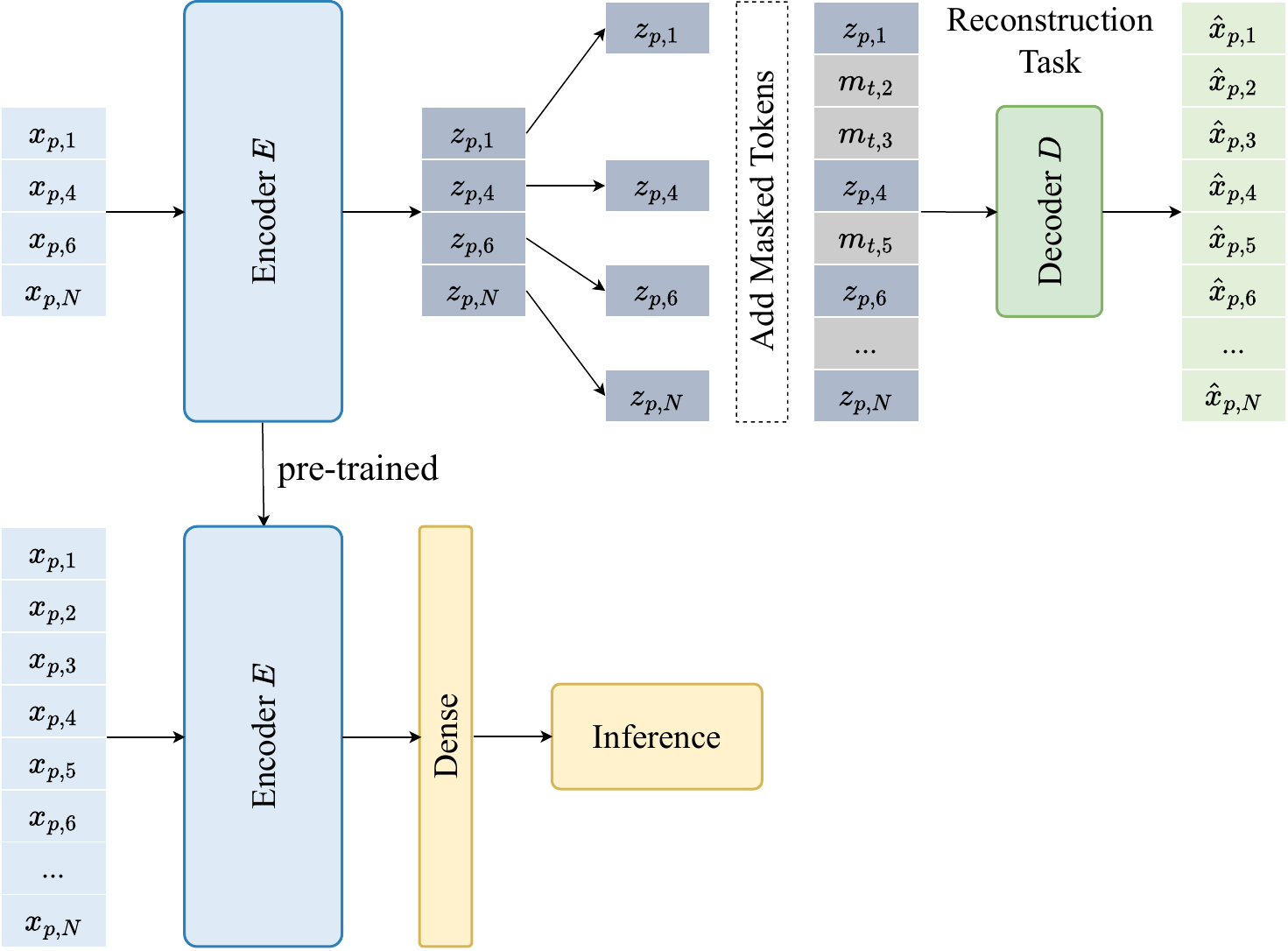}
{MAE architecture for ultrasound signals. 
The encoder processes embeddings of unmasked patches (\(x_{p,i}\)) to get latent representations (\(z_{p,i}\)). 
The decoder input combines these with mask tokens (\(m_{t,i}\)) at masked positions, where \(m_{t,i}\) represents a shared learnable embedding plus  positional information (\(p_i\)). 
Positional embeddings (\(p_i\), added before encoder and again before decoder) are omitted for clarity. 
The decoder reconstructs the full sequence (\(\hat{x}_{p,i}\)). 
Lower part: Fine-tuning uses the pre-trained encoder for downstream inference.\label{fig:vitmae}}

\subsection{Ultrasound Time-Signals vs. Images}\label{sec:ultrasound-vs-images}
As described in Section~\ref{sec:introduction}, images and ultrasound signals share fundamental similarities, making MIM approaches reasonable for pre-training. To appropriately configure the hyperparameters for processing ultrasound signals, an analysis was conducted on the differences between ultrasound signals and images, focusing on data structure, complexity, and redundancy. These comparisons served to establish appropriate scales for the model size and optimize critical components such as the Multi-Head Self-Attention mechanism. Moreover, these comparisons guided the design of key preprocessing steps, such as patching and masking, allowing their configurations to be specifically tailored to the characteristics of ultrasound signals.

\textbf{Data Structure}
Images have a multidimensional data structure of $x\in \mathbb{R}^{H\times W\times C}$, where $H$ and $W$ represent the image resolution (height and width), and $C$ is the number of color channels. Before processing using the ViT, this structure must be adapted to create a one-dimensional sequence of vectors with $x_p\in \mathbb{R}^{N\times (P^2C)}$. This is achieved by dividing the image into $N$ flattened patches of size $P^2C$ \cite{dosovitskiy_image_2021}. During this process, direct local relationships between pixels within patches and across patch boundaries are lost. Although positional embeddings are used to encode the position of each patch within the image, the internal two-dimensional pixel structure and color relationships within each patch is not directly taken into account. This means that the model must reconstruct these relationships during training, relying on mechanisms like Self-Attention to effectively capture both local and global dependencies, which demands substantial capacity.

Compared with images, the structure of ultrasound signals is far simpler, as they are one-dimensional time  series without color channels. With a constant sampling rate, the time axis can be treated analogously to the spatial dimension. From this perspective, an ultrasound signal can be considered equivalent to a one-dimensional grayscale image of the form \mbox{$x\in \mathbb{R}^{1\times W\times 1}$}, where the signal amplitude corresponds to pixel intensity values along a single spatial dimension $W$. As the signal is already a one-dimensional sequence, it can be directly divided into one-dimensional patches of $x_p\in \mathbb{R}^{N\times P}$ without disrupting the internal structure within each patch, as shown in Fig.~\ref{fig:patches}.

Specifically, a signal $x \in \mathbb{R}^{L}$ is reshaped into a sequence of $N=L/P$ patches $x_p \in \mathbb{R}^{N \times P}$. Each patch $x_{p,i} \in \mathbb{R}^P$ is then linearly projected into the model embedding space $\mathbb{R}^{d_{\text{model}}}$ using a matrix $W_{\text{proj}} \in \mathbb{R}^{P \times d_{\text{model}}}$. A learnable positional embedding $p_i \in \mathbb{R}^{d_{\text{model}}}$ is added to each projected patch embedding before feeding the sequence into the transformer encoder. 

Consequently, the preprocessing step of flattening, which is required for image patches, is unnecessary for signal patches, and was therefore not implemented in our methodology. Furthermore, because the model does not need to reconstruct disrupted local relationships, smaller models may be sufficient, as discussed in the next section on data complexity.

\Figure[t!](topskip=0pt, botskip=0pt, midskip=0pt)[width=\textwidth]{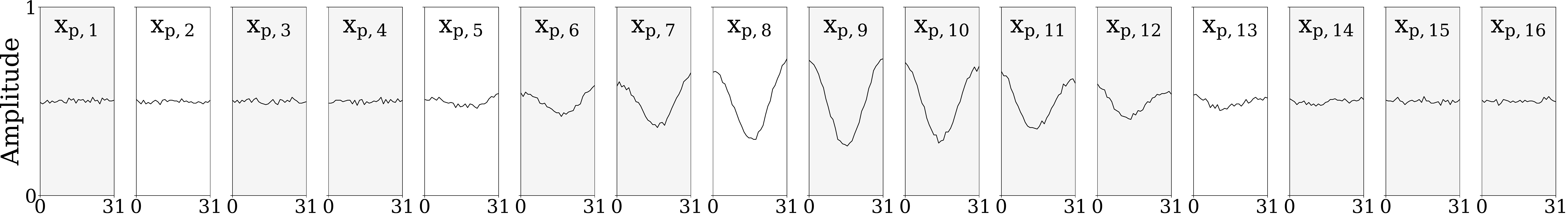}
{Process of dividing a one-dimensional ultrasound signal into non-overlapping patches.\label{fig:patches}}

\textbf{Data Complexity}
Data with complex patterns and many details, such as images or speech, typically require high-capacity networks to build robust representations, as they contain a high density of information across many features. In contrast, ultrasound signals often have periodic structures and can be described by a smaller set of features, such as frequency, amplitude, phase, and signal-to-noise ratio (SNR), which can vary significantly. When considering information density as an indicator of data complexity, Shannon entropy can be used to quantitatively compare the complexities of images and ultrasound signals. However, it is important to recognize that Shannon entropy does not fully capture the structural complexity of the data, such as spatial correlations or patterns.

To quantify and compare the statistical complexity of ultrasound  signals and image data, we computed Shannon entropy for both our dataset of ultrasound signals (\ref{sec:datasets}) and the widely used CIFAR-100 \cite{krizhevsky_learning_2009} dataset. Given that both datasets have an 8-bit resolution, the Shannon entropy $H(X)$ for a discrete random variable $X$, representing either pixel intensities (for images) or amplitude values (for ultrasound signals), is defined as:
\begin{equation} \label{eq:entropy}
H(X) = - \sum_{x=0}^{255} p(x) \log_2(p(x)),
\end{equation}
where $p(x)$ represents the probability of observing a specific value $x$ in the range $[0, 255]$.

For the CIFAR-100 dataset, which contains color images, entropy is calculated independently for each color channel (red, green, and blue) based on the pixel intensity distributions. The overall entropy of the dataset is then determined as the mean of the entropy values of the three channels:
\begin{equation}
H_{\mathrm{CIFAR}} = \frac{H_{\mathrm{red}} + H_{\mathrm{green}} + H_{\mathrm{blue}}}{3} \approx 6.88~\mathrm{bits}.
\end{equation}
For ultrasound signals, the entropy is computed using the amplitude distributions across the entire dataset, resulting in a significantly lower entropy of:
\begin{equation}
H_{\mathrm{ultrasound}} \approx 4.61~\mathrm{bits}.
\end{equation}

The lower Shannon entropy of ultrasound signals indicates significantly simpler data characteristics compared to images, suggesting that models with a much lower capacity are sufficient to learn effective representations. Based on this observation, we employed substantially smaller models than those used in the vision transformers described by Dosovitskiy et al. \cite{dosovitskiy_image_2021}. While their ViT-Base and ViT-Large architectures comprise approximately 86M and 307M parameters, respectively, the encoder part of our models range from 17K parameters in the Tiny ($T$) variant to 2.7M in the Large ($L$) variant (Section~\ref{sec:model-size}).This was achieved by experimenting with model dimensionalities ranging from $d_\mathrm{model}=32$ to $192$, with up to $h=6$ attention heads and a maximum of $l=9$ layers in the Large ($L$) variant, as shown in Table~\ref{tab:model-size}.

\textbf{Patch Size}
An important parameter for achieving high quality representations is the patch size, which we analyzed together with the masking ratio. The use of patches is crucial for enabling vision transformers to process high-resolution images with manageable computational effort. The size of the patches represents a tradeoff between computational resources and accuracy \cite{dosovitskiy_image_2021, beyer_flexivit_2023, xie_simmim_2022, hu_exploring_2022}. Smaller patches generally improve accuracy, especially in image classification, but also result in longer input sequences, leading to a quadratic increase in the computational costs for transformers \cite{vaswani_attention_2017}. Other studies have suggested that larger patch sizes can be beneficial in specific scenarios, such as tumor segmentation \cite{mojtahedi_towards_2022}. In our study, we analyzed four different patch sizes for each masking ratio, ranging from $P = 8$~samples as the smallest to $P=128$~samples as the largest (Section~\ref{sec:patching-masking}). 

These relatively small patch sizes stem from the fixed signal length of $W = 512$~samples, which we selected because it is sufficient for most ultrasound applications but significantly smaller than the number of pixels per image in datasets such as CIFAR-100 \cite{krizhevsky_learning_2009} or ImageNet \cite{deng_imagenet_2009}. Consequently, this results in input vectors with only a few dimensions $d_\mathrm{input}$, corresponding to the number of samples in each patch. This relatively low dimensionality necessitates a substantial increase during the linear projection to the model dimensionality $d_\mathrm{model}$, particularly to create a feature space with a sufficient representation capacity. For example, a patch size of $P = 16$ leads to input vectors with $d_\mathrm{input} = 16$ dimensions. When projected linearly to a model dimensionality of $d_\mathrm{model} = 128$, as in our model $M$, it represents an eightfold expansion of the embedding space. In comparison, the ViT architecture by Dosovitskiy et al. \cite{dosovitskiy_image_2021} used patch sizes of $16 \times 16 \times 3$ pixels for images, yielding vectors with $d_\mathrm{input} = 768$ dimensions, which directly align with the model dimensionality of $d_\mathrm{model} = 768$ in ViT-Base.

This difference in vector space expansion for ultrasound signals, compared to ViT for images could potentially lead to unstable training dynamics. To address this, we utilized smaller model dimensionalities than those employed in ViT, which is a feasible approach given the lower complexity of ultrasound signals relative to images, as described in the previous \textit{Data Complexity} paragraph. However, low values for $d_\mathrm{model}$ directly affect the dimensions allocated to the individual attention heads. In standard transformer implementations, the dimensionality of each attention head is defined as $d_\mathrm{head} = d_\mathrm{model} / h$, where $h$ is the number of attention heads. With reduced model dimensionalities tailored to the simpler nature of ultrasound signals, this division further limits the size of the subvector spaces within each attention head, potentially limiting their ability to capture diverse aspects of the data. To mitigate this, we explored the use of non-square query, key, and value weight matrices, offering greater flexibility in subvector space dimensionality while maintaining the overall dimensionality of $d_\mathrm{model}$ (Section~\ref{sec:attention-head-optimization}).

\textbf{Masking Ratio}
The choice of an appropriate masking ratio is important for the effectiveness of representation learning. The masking ratio determines the proportion of input data that is hidden during training, requiring the model to reconstruct the missing parts. If the masking ratio is too low, the reconstruction task becomes trivial for the model, as it can easily infer the missing data from the visible information. This may lead the model to rely on simple heuristics, such as interpolation, rather than learning meaningful and general representations. If the masking ratio is too high, the task may become overly challenging, as the model might lack sufficient context to reconstruct the input.

The ideal masking ratio is hypothesized to correlate with the redundancy of the input data. For instance, in BERT \cite{devlin_bert_2019}, the high information density of natural language supports effective representation learning with a masking ratio of only $15\%$. In contrast, images that contain more redundant information, tend to perform better with masking ratios of approximately $75\%$ \cite{he_masked_2022}. For videos, where spatiotemporal properties introduce even greater redundancy, masking ratios as high as $90\%$ have been shown to perform well \cite{feichtenhofer_masked_2022}.

In the case of ultrasound signals, it was assumed that the data exhibits high redundancy owing to its periodic characteristics and extended segments containing only noise, particularly in signals with short tone bursts. Consequently, three high masking ratios were evaluated: $62.5\%$, $75\%$, and $87.5\%$ (Section~\ref{sec:patching-masking}).

\subsection{Datasets}\label{sec:datasets}
In our experiments, we used two different datasets: the first consists entirely of \textit{synthetically generated} ultrasound signals, while the second contains \textit{measured} ultrasound signals. In the first step, synthetic data were used to optimize the fundamental hyperparameters, as they can be generated in large quantities with minimal effort (Sections~\ref{sec:model-size}, \ref{sec:attention-head-optimization} and \ref{sec:patching-masking}). In the second step, we used measured ultrasound signals from our experimental setup to evaluate how well the model’s learned representations generalize to a real-world application. (Section~\ref{sec:real-signals}). 

Table~\ref{tab:datasets} presents a comparison of the synthetic and measured datasets. The synthetic dataset is five times larger but shares the same sampling rate and a signal length of $N=512$~samples, which is suitable for the majority of ultrasound applications. To make the downstream task more challenging, a low resolution of 8~bits was selected for both pre-training and fine-tuning. Additionally, the low resolution is beneficial for reducing hardware requirements, offering a practical advantage for industrial measurement systems where cost efficiency and real-time processing are critical.

\begin{table}[h]
\centering
\caption{Comparison of synthetic and measured datasets, highlighting differences in size and key signal attributes.}
\begin{tabularx}{\columnwidth} {l Y Y Y Y l}
    \toprule
    & \multicolumn{2}{c}{\textbf{Synthetic Data}}
    & \multicolumn{2}{c}{\textbf{Measured Data}}
    & \multicolumn{1}{l}{\textbf{Unit}} \\
    \midrule
    Size (train/val/test) & \multicolumn{2}{c}{48K/6K/6K} & \multicolumn{2}{c}{9.6K/1.2K/1.2K} & \\
    Signal length & \multicolumn{2}{c}{512} & \multicolumn{2}{c}{512} & Samples\\
    Sampling rate & \multicolumn{2}{c}{60} & \multicolumn{2}{c}{60} & MHz\\
    Resolution & \multicolumn{2}{c}{8} & \multicolumn{2}{c}{8} & bit\\
    ToF Classes & \multicolumn{2}{c}{200} & \multicolumn{2}{c}{200} & \\
    \midrule
    & min & max & min & max & \\ 
    Signal frequency & 1.0 & 4.0 & 2.0 & 2.4 & MHz \\
    Samples per period & 15 & 60 & 25 & 30 & Samples \\
    Amplitude (norm.) & 0.2 & 1 & 0.8 & 1 & V \\
    Burst length & 200 & 400 & 75 & 180 & Samples \\
    Peak SNR & 18 & 38 & 27 & 29 & dB \\
    \bottomrule
\end{tabularx}
\label{tab:datasets}
\end{table}

The primary distinction between the two datasets is the variability of key attributes, including signal frequency, amplitude, burst length, and peak SNR. The synthetic dataset exhibits a high degree of variability in these attributes, which is advantageous for learning generalizable signal representations during the reconstruction task. In contrast, the measured dataset demonstrates significantly reduced variability. This reduced variability can be attributed to the transducer's limited bandwidth and inherent limitations of the measurement setup, which naturally constrain the range of signal characteristics, such as frequency and amplitude. This design aligns with our objective of evaluating whether representation learning from a large and diverse synthetic dataset can effectively support real-world downstream tasks.

An important aspect is the burst length of the measured signals, which is significantly shorter than that of the synthetic data (Table \ref{tab:datasets}). This intentional discrepancy aims to evaluate the ability of the learned representations from pre-training to generalize to signals with characteristics that significantly diverge from the trained range.

\subsection{Downstream Task}
The selected downstream task for fine-tuning the models after pre-training was the determination of the absolute time-of-flight (ToF) of ultrasound tone burst signals. This task is intended to evaluate the quality of the learned representations as a substantial challenge for the model. Simultaneously, it ensures practical relevance, demonstrating that the representations, particularly those learned from synthetic data, are robust and applicable to real-world scenarios.

Time-of-flight estimation, including distance measurement and material inspection, is a fundamental task in ultrasonic applications. While \textit{relative} ToF shifts can often be determined using simple methods, such as detecting zero crossings after a defined threshold, accurately determining the \textit{absolute} ToF for signals is significantly more challenging. This difficulty arises from the fact that the exact start of the tone burst is often obscured by noise and, as a result, becomes blurred, making it challenging to identify. This issue is exacerbated when, as in our case, the signals exhibit significant variability in characteristics, such as amplitude, frequency, number of cycles, and SNR. In this scenario, a fixed offset from the phase crossings to the signal onset does not exist and therefore cannot be used to determine the absolute ToF.

Specifically, the ToF downstream task is framed as a classification problem in which the model is used to identify the first sample that marks the onset of the tone burst. This is particularly challenging for signals with low amplitudes and low SNR because the start of the burst is often heavily obscured. Figure \ref{fig:tof} provides an example of this classification task, showing two signals from the synthetic dataset that exhibit significant differences in their characteristics, including the frequency, amplitude, number of cycles, and their respective ToF labels. The initial sample can occur anywhere within an interval of 0-199, corresponding to a total of 200 classes.

\Figure[t!](topskip=0pt, botskip=0pt, midskip=0pt)[width=0.999\columnwidth]{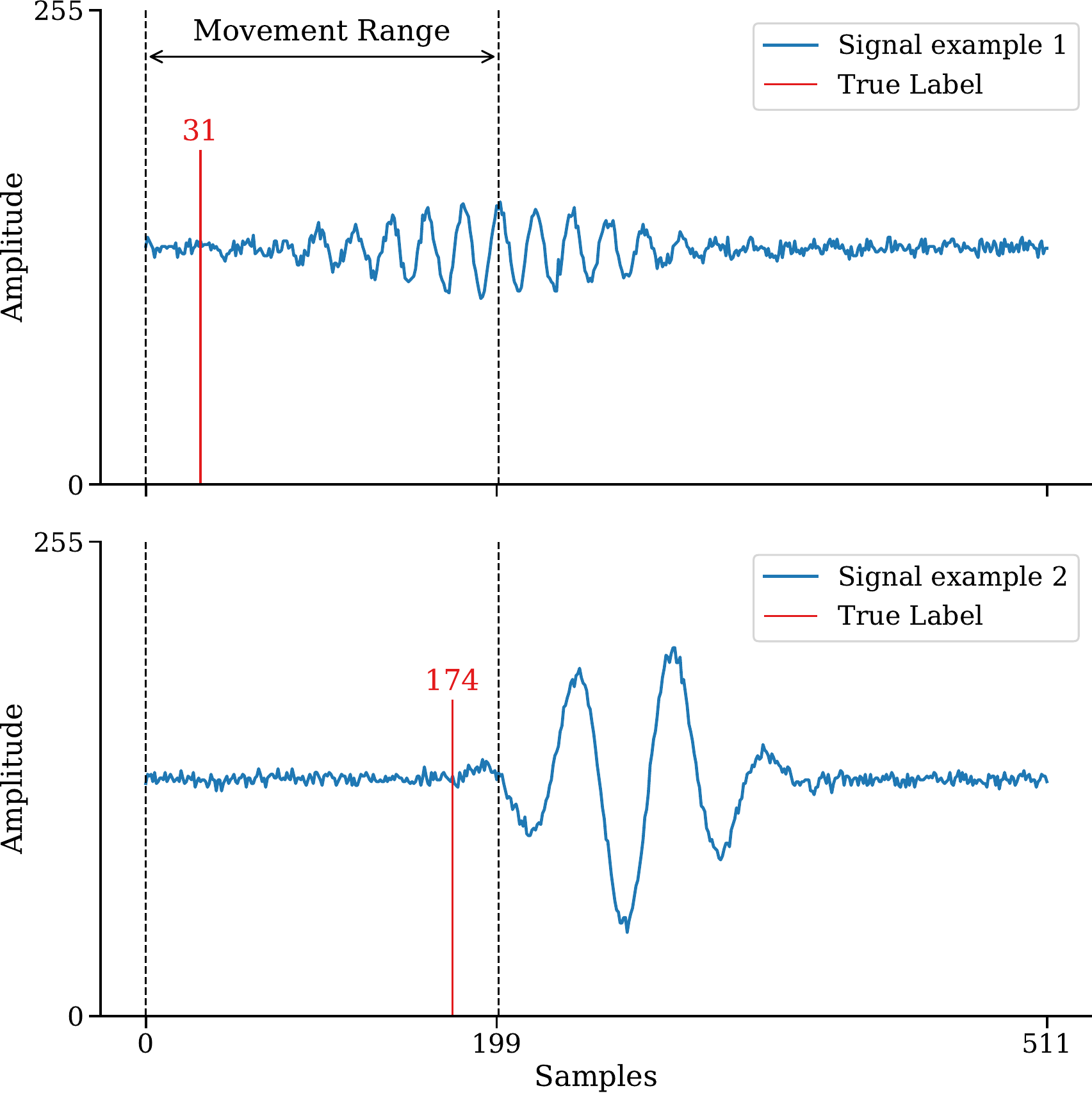}
{Two examples of synthetic ultrasound signals with ToF labels. The figure highlights variations in frequency, amplitude, and signal noise, showcasing the challenges of classifying the precise onset of tone bursts in diverse signal conditions.\label{fig:tof}}

Label generation for synthetic signals is inherently straightforward because the start of the burst is predetermined during the signal creation process, ensuring that each signal is assigned to a valid ToF label. In contrast, measured signals require a separate process for label determination. In our case, cross-correlation was identified as the most reliable method for this purpose, as it acts as a matched filter. This approach requires the specific excitation signal $s(t)$ associated with each individual received signal $r(t)$ to accurately determine the ToF label. The cross-correlation $C(\tau)$ between the measured signal and its corresponding excitation signal is given by
\begin{equation}
C(\tau) = \sum_{t=0}^{N-1} r(t) \cdot s(t + \tau),
\end{equation}
where $\tau$ is the time lag, $N$ is the length of the signals, and $t$ is the sample index. The position of maximum $C(\tau)$ indicates the time lag $\tau_{\text{max}}$, at which the signals are best aligned. This time lag served as the ToF label, representing the first sample of the tone burst in the measured signal.

In contrast, the transformer model must predict the ToF labels without any access to the excitation signal. This represents a significant challenge, as the model must infer the label directly from the received signal, despite variations in signal characteristics such as amplitude, frequency, and noise. Unlike the matched filter, the transformer relies solely on the learned representations of the signals to perform this task, making it a robust test of its generalization capabilities.

\subsection{Evaluation Metrics} \label{sec:metrics}
The performance of the model during pre-training was evaluated using the mean absolute error (MAE), calculated over the reconstructed masked samples according to the objective function defined in Eq.~\eqref{eq:mae_loss}. Lower MAE values indicate higher reconstruction quality, with zero representing a perfect reconstruction. However, owing to the inherent noise in ultrasound signals, a perfect reconstruction is not achievable in practice.

Importantly, a low MAE does not necessarily imply that the learned representations are meaningful for the downstream tasks. For example, when using a low masking ratio, the reconstruction task becomes trivial, and the model may simply interpolate local structures rather than learning generalizable features. To assess the effectiveness of the learned representations, we therefore evaluated their performance on a downstream classification task.

Depending on the dataset and signal characteristics, we report the \mbox{top-1} accuracy in combination with either \mbox{top-2} or \mbox{top-5} accuracy to appropriately capture the classification performance. While the top-1 accuracy reflects the exact prediction of the correct ToF class, the \mbox{top-2} and \mbox{top-5} accuracy tolerate small deviations and indicate whether the correct class lies among the closest alternatives, providing a more nuanced view of the model's temporal precision.

In addition, we report a classification-derived MAE in nanoseconds computed by mapping each predicted class index to its corresponding ToF value. Although the model was not trained for regression, this metric provides a domain relevant measure of temporal accuracy.

All reported metrics are averaged across ten independent training runs and presented with their respective standard deviations to indicate performance stability.

\section{Experiments and Results}
\subsection{Model Size}\label{sec:model-size}
Our first goal was to determine an appropriate model size that could effectively learn a robust representation of ultrasound signals while achieving optimal accuracy on the downstream task. The model size is primarily influenced by the architecture's hyperparameters, including the model dimensions $d_{\mathrm{model}}$, attention head dimensions $d_{\mathrm{head}}$, number of attention heads $h$, and number of layers $l$ (encoder and decoder blocks). We compared four different model configurations ($T\rightarrow\text{tiny}$, $S\rightarrow\text{small}$, $M\rightarrow\text{medium}$ and $L\rightarrow\text{large}$), each with substantial variations in size, achieved by simultaneously increasing $d_{\mathrm{model}}$, $h$, and $l$, as listed in Table~\ref{tab:model-size}. The attention head dimensions were set according to the standard practice of $d_{\mathrm{head}} = d_{\mathrm{model}}/h$.

\begin{table}[ht]
\centering
\caption{Configurations of transformer models with varying sizes to systematically investigate the impact of model capacity on performance.}
\begin{tabularx}{\columnwidth} {l Y Y Y Y Y Y Y Y}
    \toprule
    & \multicolumn{2}{c}{\textbf{Model-Dim.}}
    & \multicolumn{2}{c}{\textbf{Att. Heads}}
    & \multicolumn{2}{c}{\textbf{Layer}}
    & \multicolumn{2}{c}{\textbf{Parameter}}\\   
    & \multicolumn{2}{c}{$d_{\mathrm{model}}$}
    & \multicolumn{2}{c}{$h$}
    & \multicolumn{2}{c}{$l$}
    & \multicolumn{2}{c}{$W$}\\
    & ENC & DEC & ENC & DEC & ENC & DEC & ENC & DEC \\ 
    \midrule
    T & 32 & 16 & 1 & 1 & 2 & 1 & 17K & 2K \\
    S & 64 & 32 & 2 & 2 & 3 & 1 & 100K & 9K  \\
    M & 128 & 64 & 4 & 4 & 6 & 2 & 795K & 67K  \\
    L & 192 & 96 & 6 & 6 & 9 & 3 & 2.7M & 225K  \\
    \bottomrule
\end{tabularx}
\label{tab:model-size}
\end{table}

Table~\ref{tab:model-size-result} shows the performance of the different model sizes on the downstream task with and without pre-training. Notably, no additional or alternative training data were used for the pre-trained models. To ensure a fair comparison, hyperparameters including learning rate, batch size, and dropout rate, were kept consistent across experiments (Section~\ref{sec:implementation-details}). For model sizes $S$, $M$, and $L$, \textit{fine-tuning} with pre-trained weights consistently outperforms training \textit{from scratch} after 200 epochs. However, for model $T$, training from scratch yields better results under the same conditions. The performance gap in favor of the pre-trained models is particularly evident in the learning curves in Fig.~\ref{fig:model-size-combined}, where models $S, M$ and $L$ achieve significantly higher \mbox{top-1} and \mbox{top-5} accuracy after only a few epochs. Only for model $T$ do randomly initialized weights outperform pre-trained weights, likely because the model's limited capacity prevents it from learning meaningful representations during pre-training, which may even hinder fine-tuning by starting from a suboptimal parameter space. Compared to MAE in the field of image processing \cite{he_masked_2022}, the improvement owing to pre-training is significantly higher for ultrasound signals. One potential explanation for this phenomenon is that the one-dimensional nature of these signals makes learning meaningful representations much easier. Furthermore, because flattening is unnecessary for one-dimensional signals, the internal structure within each patch remains intact, avoiding the disruption typically caused by this process, as discussed in the \textit{Data Structure} paragraph in Section~\ref{sec:ultrasound-vs-images}

\begin{table}[ht]
\centering
\caption{Performance metrics for pre-trained versus from-scratch models, showing consistent improvements in accuracy through self-supervised pre-training for all model sizes.}
\begin{tabularx}{\columnwidth} { l c Y Y Y Y}
    \toprule
    & \textbf{MAE}
    & \multicolumn{2}{c}{\textbf{Top-1 Accuracy}}
    & \multicolumn{2}{c}{\textbf{Top-5 Accuracy}} \\
    & pre-tr. & scratch & fine-tuning & scratch & fine-tuning \\  
    \midrule
    T   &  3.04 & 22.15  {\tiny$\!\pm$ \!7.60} & 14.59 {\tiny$\!\pm$ \!7.88} & 70.58  {\tiny$\!\pm$ \!11.95}  & 50.77 {\tiny$\!\pm$ \!22.94}\\
    S  &  1.89 & \textbf{45.81} {\tiny$\!\pm$ \!15.87} & 62.19 {\tiny$\!\pm$ \!6.12} & \textbf{88.76}  {\tiny$\!\pm$ \!11.69} & 79.35  {\tiny$\!\pm$ \!2.09}\\
    M &  \textbf{1.38} & 44.25 {\tiny$\!\pm$ \!18.81} & \textbf{75.38} {\tiny$\!\pm$ \!9.24} & 85.11  {\tiny$\!\pm$ \!17.04}  & \textbf{98.98} {\tiny$\!\pm$ \!1.31}\\
    L  &  \textbf{1.38} & 45.45 {\tiny$\!\pm$ \!19.84} & 74.45 {\tiny$\!\pm$ \!4.00} & 83.69  {\tiny$\!\pm$ \!25.43}  & 98.96 {\tiny$\!\pm$ \!0.34}\\
    \bottomrule
\end{tabularx}
\label{tab:model-size-result}
\end{table}

\Figure[t!](topskip=0pt, botskip=0pt, midskip=0pt)[width=0.999\columnwidth]{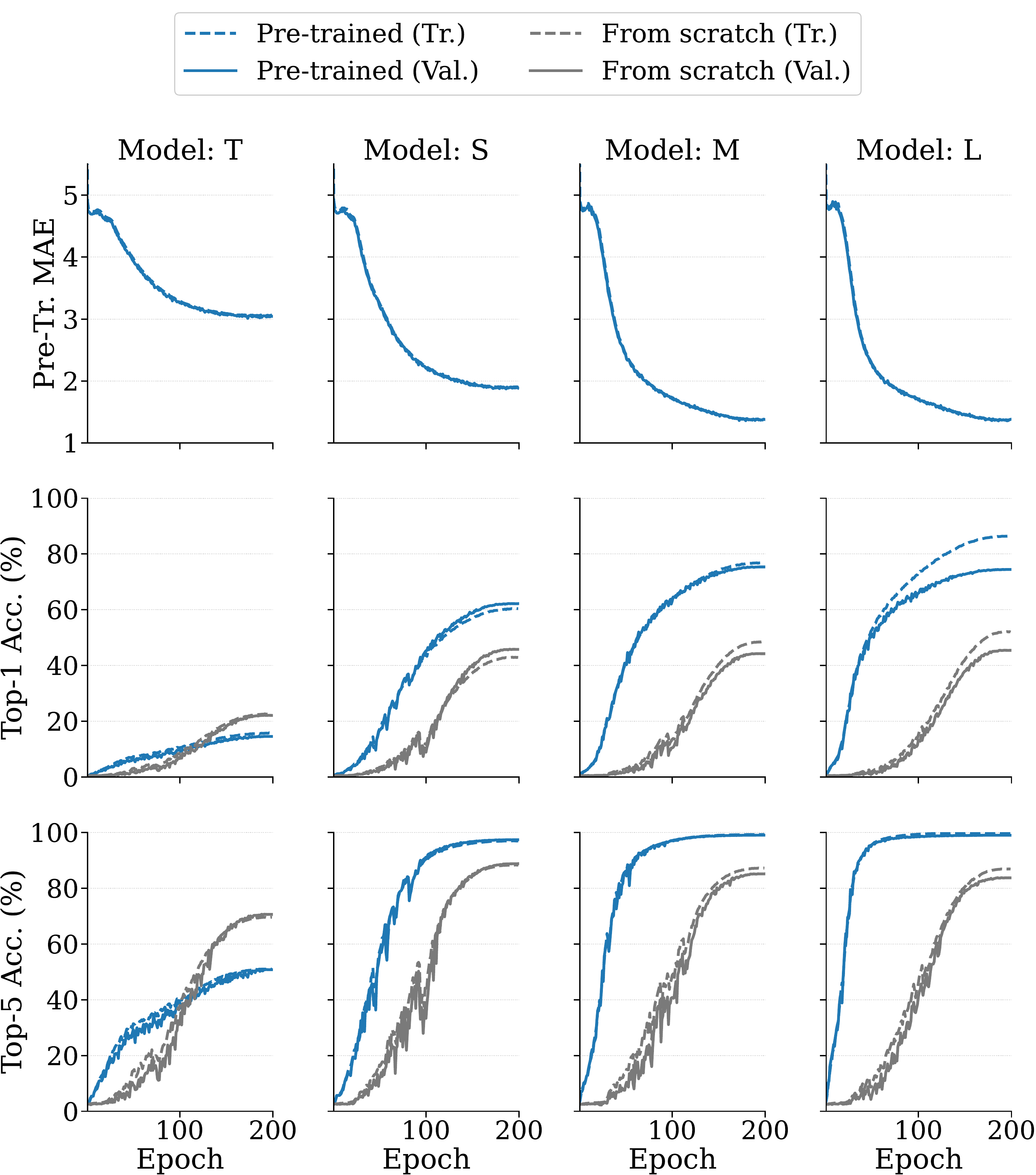}
{Training and validation curves comparing pre-trained and from-scratch models across different model size configurations. Pre-trained models demonstrate faster convergence and higher accuracy while avoiding overfitting, particularly in larger model configurations.\label{fig:model-size-combined}}

It is widely acknowledged that high-capacity transformer models can achieve higher accuracy than their smaller counterparts, an observation that is also present in our experimental results. However, this advantage comes at the cost of an increased risk of overfitting, particularly when the size of the training dataset is limited. Figure~\ref{fig:model-size-combined} highlights this for Model $L$, showing the gap between the training and validation accuracy, which underscores the degree of overfitting in both the from-scratch training and fine-tuning scenarios. In contrast, Model $S$ exhibits only slight overfitting, which is further mitigated by pre-training, whereas in the case of Model $M$, overfitting is fully eliminated through pre-training.

An additional consideration regarding model size is that the mean absolute error (MAE) metric in Table~\ref{tab:model-size-result} and Fig.~\ref{fig:model-size-combined} shows significantly better signal reconstruction performance for the $M$ and $L$ models compared with the smaller $T$ and $S$ variants. This suggests that $T$ and $S$ lack sufficient capacity to effectively learn robust signal representations. Furthermore, increasing the model capacity beyond $M$ does not yield any noticeable improvement in the MAE metric or downstream task accuracy.

As hypothesized in (Section~\ref{sec:patching-masking}) our comparison of different model sizes demonstrates that significantly smaller models are sufficient for ultrasound signals compared with ViT models designed for images. This compact size is particularly advantageous in industrial settings because it enables deployment on edge devices with limited computational resources, enabling cost-effective and efficient real-time signal analysis.

\subsection{Attention-Head Optimization} \label{sec:attention-head-optimization}
Using multiple attention heads allows the model to learn a diversified representation, with each attention head focusing on different aspects of the input data. To achieve this, each attention head operates within its own subvector space of dimensionality $d_{\mathrm{head}}$. By default, the dimensionality of these subvector spaces is set to $d_{\mathrm{head}} = d_{\mathrm{model}}/h$, making $d_{\mathrm{head}}$ significantly smaller than $d_{\mathrm{model}}$ to maintain the computational efficiency.  Consequently, $d_{\mathrm{head}} = 32$ for all model sizes ($T$, $S$, $M$, and $L$) from Section~\ref{sec:model-size}. This relatively small feature space stems from the smaller size of signal patches compared to image patches, despite the linear projection of patches into a larger vector space $d_{\mathrm{model}}$, as described in the \textit{Patch Size} paragraph in Section~\ref{sec:ultrasound-vs-images}. To prevent the feature spaces within the attention heads from becoming too small, we explored the use of non-square query, key, and value weight matrices, defined as 
\begin{equation} \label{eq:weights}
\begin{split}
W_Q, W_K, W_V &\in \mathbb{R}^{d_\mathrm{model} \times h d_\mathrm{head}} \\
\text{where } d_\mathrm{head} &\neq \frac{d_\mathrm{model}}{h}.
\end{split}
\end{equation}
This is in contrast to standard implementations, which typically set these matrices to the dimensions of $d_\mathrm{model} \times d_\mathrm{model}$. The output projection matrix is defined as $W_O \in \mathbb{R}^{h d_\mathrm{head} \times d_\mathrm{model}}$, ensuring that $d_\mathrm{model}$ is preserved across layers.

Table~\ref{tab:head-dimension} shows that the \mbox{M-dh64} model, with a head dimensionality of $d_\mathrm{head} = 64$ and $h = 4$ attention heads, achieves significantly higher \mbox{top-1} and \mbox{top-5} accuracies than the \mbox{M-dh32} model, which uses the standard square matrices and a head dimensionality of $d_\mathrm{head} = 32$. Furthermore, the MAE metric decreased from $1.38$ to $1.33$, suggesting improved representation learning. However, increasing $d_\mathrm{head}$ to $128$ in the \mbox{M-dh128} model does not result in any additional performance gains and leads to overfitting.

\begin{table}[ht]
\centering
\caption{Analysis of attention head dimensionality and its effects on model performance and parameter count.}
\begin{tabularx}{\columnwidth} {l c Y Y c c}
    \toprule
    & \textbf{Pre-tr}
    & \multicolumn{2}{c}{\textbf{Downstream Accuracy}}
    & \multicolumn{2}{c}{\textbf{Parameter} $W$} \\
    & MAE & Top-1 & Top-5 & ENC & DEC \\
    \midrule
    M-dh32  &  1.38 & 75.38 {\tiny$\!\pm$ \!9.24} & 98.95 {\tiny$\!\pm$ \!1.30} & 795K & 67K\\
    M-dh64 (base)  &  1.33 & \textbf{79.28} {\tiny$\!\pm$ \!1.80} & 99.46 {\tiny$\!\pm$ \!0.26} & 1.2M & 100K\\
    M-dh128 &  1.40 & 74.87 {\tiny$\!\pm$ \!7.73} & 99.00 {\tiny$\!\pm$ \!0.89} & 2M & 167K\\
    \midrule
    M-dh64-h3  &  \textbf{1.32} & 78.70 {\tiny$\!\pm$ \!1.59} & \textbf{99.54} {\tiny$\!\pm$ \!0.13} & 992K & 84K\\
    M-dh64-h5  &  1.46 & 76.90 {\tiny$\!\pm$ \!4.20} & 99.28 {\tiny$\!\pm$ \!0.21} & 1.4M & 117K\\
   \bottomrule
\end{tabularx}
\label{tab:head-dimension}
\end{table}

Reducing the number of attention heads from $d_\mathrm{head} = 4$ to $d_\mathrm{head} = 3$ yielded comparable results but with slightly more unstable learning curves in the downstream application. Conversely, increasing $d_\mathrm{head} = 5$ worsened both the MAE metric and the the \mbox{top-1} and \mbox{top-5} accuracies and leads to significantly more unstable downstream learning curves. Overall, the best results were obtained with the \mbox{M-dh64} model, which was therefore selected as the \textit{base} model for further experiments.

\subsection{Patching and Masking}  \label{sec:patching-masking}
In conjunction with the variation in patch size, we also experimented with three different masking ratios, $62.5\%$, $75\%$, and $87.5\%$, to evaluate their impact on representation learning. Using high masking ratios ensures that the reconstruction task remains sufficiently challenging even for the relatively simple and periodic structure of ultrasound signals.

Table~\ref{tab:patching} and Fig.~\ref{fig:patching} provides an overview of the pre-training and downstream task results for different combinations of masking ratios and patch sizes, including the corresponding counts of the unmasked and masked patches for a fixed signal length of $N=512$~samples. These configurations were tested using our best-performing model, M-base (\mbox{M-dh64}).

\begin{table}[ht]
\centering
\caption{Evaluation of how different masking ratios and patch sizes influence model performance.}
\begin{tabularx}{\columnwidth} {c c c c Y Y}
    \toprule
    \multicolumn{3}{c}{\textbf{M-base configuration}}
    & \textbf{Pre-tr.}
    & \multicolumn{2}{c}{\textbf{Downstream Accuracy}}\\       
    Masking & PS & m/u & MAE & Top-1 & Top-5  \\
    \midrule
    62,5\% & 8 &  40/24 & \textbf{1.00} & 47.68 {\tiny$\!\pm$ \!9.00} & 95.03 {\tiny$\!\pm$ \!2.81}\\
    & 16 &  20/12 &  1.06 & 76.45 {\tiny$\!\pm$ \!1.91} & 99.19 {\tiny$\!\pm$ \!0.25}\\
    & 32 &  10/6 &  1.36 & 82.01 {\tiny$\!\pm$ \!2.23} & 99.51 {\tiny$\!\pm$ \!0.13}\\
    & 64 &  5/3 &  1.85 & 80.57 {\tiny$\!\pm$ \!2.10} & 99.35 {\tiny$\!\pm$ \!0.30}\\
    \midrule
    75\% & 8 &  48/16  &  1.16 & 50.69 {\tiny$\!\pm$ \!7.79} & 96.11 {\tiny$\!\pm$ \!1.91}\\
    & 16 &  24/8  &  1.33 & 79.28 {\tiny$\!\pm$ \!1.80} & 99.46 {\tiny$\!\pm$ \!0.26}\\
    & 32 &  12/4  &  1.67 & \textbf{89.05} {\tiny$\!\pm$ \!2.20} & \textbf{99.86} {\tiny$\!\pm$ \!0.10}\\
    & 64 &  6/2 &  2.45 & 86.27 {\tiny$\!\pm$ \!1.57} & 99.71 {\tiny$\!\pm$ \!0.09}\\
    \midrule
    87.5\% & 8 &  56/8  &  1.86 & 58.26 {\tiny$\!\pm$ \!1.59} & 97.13 {\tiny$\!\pm$ \!9.26}\\
    & 16 &  28/4  &  2.22 & 77.60 {\tiny$\!\pm$ \!4.20} & 99.40 {\tiny$\!\pm$ \!3.75}\\
    & 32 &  14/2  &  2.72 & 76.98 {\tiny$\!\pm$ \!4.20} & 99.03 {\tiny$\!\pm$ \!0.71}\\
    & 64 &  7/1 &  3.36 & 67.45 {\tiny$\!\pm$ \!7.73} & 97.84 {\tiny$\!\pm$ \!0.09}\\
   \bottomrule
\end{tabularx}
\label{tab:patching}
\end{table}

When examining the MAE metric, it becomes evident that smaller patch sizes consistently result in more accurate reconstructions, regardless of the masking ratio. Furthermore, as expected, a lower masking ratio results in reduced MAE values, as the reconstruction task becomes less challenging. When evaluating the downstream task using the \mbox{top-1} and \mbox{top-5} metrics, the best performance was achieved with a patch size of $P=32$ and a masking ratio of $75\%$. In contrast, the smallest patch size of $P=8$ showed the lowest accuracy, highlighting a difference between reconstruction quality and the effectiveness of the learned representations for downstream applications. Larger patch sizes, while causing a slight reduction in downstream accuracy, compared to the optimal patch size, still significantly outperformed the smallest patch size of $P=8$.

\Figure[t!](topskip=0pt, botskip=0pt, midskip=0pt)[width=0.999\linewidth]{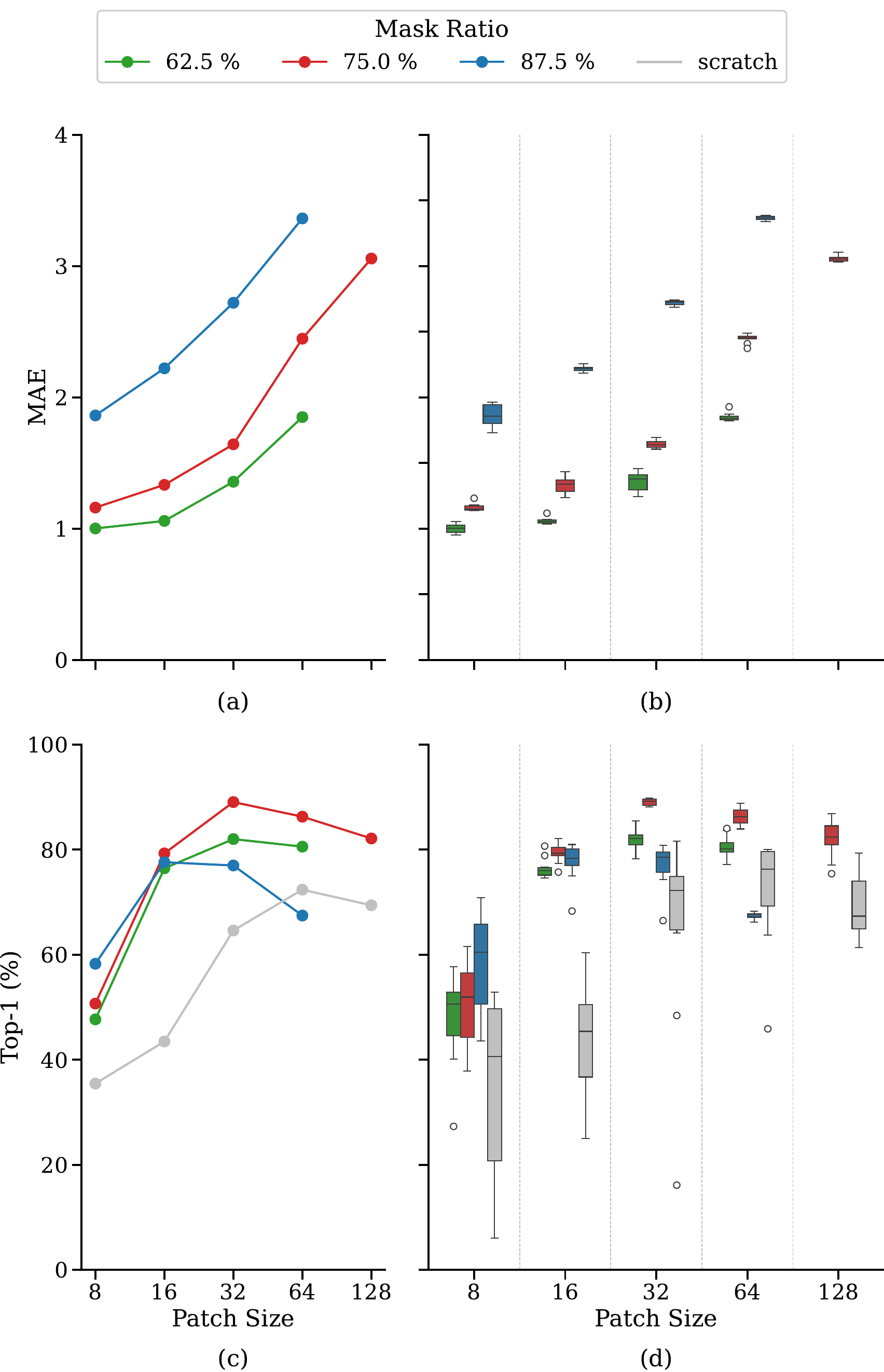}{Performance evaluation of pre-training strategies (mask ratio and scratch vs. patch size) over 10 runs. The top row (a, b) shows the pre-training Mean Absolute Error (MAE) reconstruction performance: (a) Mean values across runs and (b) corresponding result distribution using box plots. The bottom row (c, d) shows the downstream \mbox{top-1} performance: (c) Mean values across runs and  (d) corresponding result distribution using box plots.\label{fig:patching}}

This observation is in contrast to the findings in the field of image classification, where smaller patch sizes often lead to better downstream performance owing to their ability to capture fine-grained details and local features. Similarly, in the context of the downstream task of ToF estimation, which demands high precision and sensitivity to fine details, it is reasonable to assume that smaller patch sizes would perform better. We hypothesize that small patch sizes might be less effective for ultrasound signals because of their periodic nature, which distinguishes them significantly from images and other types of time series data. If the patch size is too small relative to the periodic characteristics of the data, the model may struggle to capture the periodic patterns effectively. This limitation arises because smaller patches may not encompass a significant portion of a single period, which, depending on the signal frequency, range from $16$ to $60$~samples (Table~\ref{tab:datasets}). As a result, periodic information is fragmented across multiple patches, requiring the model to infer these patterns from the relationships between patches. Although theoretically possible because of the Self-Attention mechanism in transformers, this process significantly increases the complexity of learning, as the model must infer periodic patterns from relationships scattered across multiple patches.

Another potential explanation for the poor accuracy observed with small patches is that the reconstruction task may sometimes be addressed through simple interpolation between visible patches. This is particularly the case for small patch sizes, where the masked sections between them are very small. As a result, the model may prioritize solving the reconstruction task with minimal effort rather than capturing the intrinsic structure of the signal, leading to less robust representations. This hypothesis is supported by the observation that the highest masking ratio of $87.5\%$ improves the downstream performance with a small patch size of $P=8$.

As in Section~\ref{sec:model-size}, we compared the pre-trained models with those trained from scratch. Although the latter benefits from attention head optimization and appropriate patch size selection, they still lag significantly in accuracy. Pre-trained models also converge faster, with smoother learning curves, and exhibit enhanced stability, as shown by the boxplots in Fig.~\ref{fig:patching} and low standard deviations in Table~\ref{tab:patching}. Despite not using additional training data, pre-trained models consistently outperform randomly initialized  models, which display unstable training and inferior results.

\subsection{Real Signals}  \label{sec:real-signals}
\textbf{Representation Quality}
This section evaluates how well representations learned through self-supervised pre-training generalize to real-world ultrasound measurements with signal characteristics unseen during training. The measured signals differ substantially from the synthetic ones in several key aspects, particularly in their ring-down behavior, governed by the properties of the piezoceramic transducer, damping effect of the backing layer, and in burst length, which falls completely outside the range encountered during pre-training (see Section~\ref{sec:datasets}). This setup tests whether the learned representations generalize to realistic ultrasound tone bursts.

We compared four pre-training scenarios: (i) training from scratch, (ii) pre-training on measured data (12K~samples), (iii) pre-training on synthetic data (60K~samples), and (iv) pre-training on a combined dataset (72K~samples). In all the cases, fine-tuning was conducted exclusively on the measured data. Table~\ref{tab:real} presents the results of this comparison. For all the experiments in this section, we used the M-base model with a masking ratio of 75\% and a patch size of $P=32$, as this configuration yielded the best performance during our optimization studies in Sections~\ref{sec:attention-head-optimization} and~\ref{sec:patching-masking}.

\begin{table}[ht]
\centering
\caption{Comparison of model generalization capabilities of M-dh64 model in downstream tasks using measured (real) signals.}
\begin{tabularx}{\columnwidth} {c c Y Y Y}
    \toprule
    \multicolumn{2}{c}{\textbf{Dataset}}
    & \textbf{Pre-tr}
    & \multicolumn{2}{c}{\textbf{Downstream Accuracy}}\\
    Pre-tr. & Fine-tuning & MAE & Top-1 & Top-5  \\
    \midrule
    - & real & - & 68.03 {\tiny$\!\pm$ \!3.16} & 96.57 {\tiny$\!\pm$ \!1.06}\\
    real &  real &  1.23 & 78.68 {\tiny$\!\pm$ \!3.11} & \textbf{100} {\tiny$\!\pm$ \!0.00}\\
    syn &  real &  1.62 & 87.46 {\tiny$\!\pm$ \!0.30} & \textbf{100} {\tiny$\!\pm$ \!0.00}\\
    syn+real &  real &  1.55 & \textbf{87.58} {\tiny$\!\pm$ \!0.47} & \textbf{100} {\tiny$\!\pm$ \!0.00}\\
   \bottomrule
\end{tabularx}
\label{tab:real}
\end{table}

Training from scratch results in the lowest \mbox{top-1} accuracy of $68.03\%$, with a relatively high standard deviation of $3.16\%$, reflecting poor and inconsistent performance. Pre-training on the measured data increases the \mbox{top-1} accuracy significantly to $78.68\%$. Pre-training on synthetic data further improves the performance, yielding an accuracy of $87.46\%$ with a substantially reduced standard deviation of $0.30\%$, demonstrating strong generalization and stability. The best performance is achieved with combined pre-training (synthetic and measured data), achieving a \mbox{top-1} accuracy of $87.58\%$, which indicates that including real data offers only minor benefits beyond synthetic pre-training.

It is noteworthy that the MAE achieved with measured data pre-training (1.23) is lower than that with synthetic pre-training (1.62), suggesting a better reconstruction of real signals. However, this lower MAE likely results from the limited variability of the measured dataset, which makes reconstruction easier but may not reflect the learning of meaningful features for downstream tasks.

These findings underscore the advantage of synthetic pre-training, which exposes the model to a broad range of signal characteristics, such as frequency, amplitude, and burst length (see Table~\ref{tab:datasets}), and thereby enabling it to learn more generalized and robust representations of ultrasound tone bursts. In contrast, the measured dataset exhibits limited variability because it is typically tailored to specific applications and does not capture the full spectrum of potential ultrasound signals. Consequently, pre-training on such narrowly focused data may constrain the ability of the model to develop representations that generalize well beyond the task-specific signal characteristics.

For example, a training dataset encompassing a wide range of signal frequencies facilitates the model’s ability to learn the underlying concept of frequency variation, in contrast to a dataset with restricted frequency diversity. This generalized understanding can be effectively transferred across various downstream tasks, even when the model encounters signals with previously unseen frequency components.

\textbf{Reducing Downstream Data}
Beyond generalization to real signals, another key advantage of pre-training is its potential to reduce the amount of labeled data required for downstream tasks. To evaluate the efficiency of the learned representations in this regard, we systematically compared the performance of our MAE approach with two baseline models for various dataset sizes.

The right panel of Fig.~\ref{fig:real-dataset-size} shows the mean absolute error (MAE) in nanoseconds, which is derived from the classification output by mapping the predicted class index to the time-of-flight values. For a detailed explanation of this metric, see Section~\ref{sec:metrics}.

A summary of the experimental results across different dataset sizes is provided in Table~\ref{tab:real-results}, with additional visualizations in Fig.~\ref{fig:real-dataset-size}. Fig.~\ref{fig:real-dataset-size} provides a visual comparison of model performance across different dataset sizes, showing the \mbox{top-1} accuracy in the left panel and the corresponding mean absolute error (MAE) in nanoseconds in the right panel. The MAE is computed by mapping the \mbox{top-1} class predictions to their associated time-of-flight values. While the model operates as a classifier, this metric enables an intuitive domain-relevant interpretation of the prediction error.

Specifically, we utilized two different CNN architectures as reference points to contextualize the MAE performance. A naive baseline, characterized by a straightforward CNN design with a similar parameter count to our MAE model, serves to assess performance relative to a simple, commonly used CNN approach. In contrast, the strong baseline employs a more sophisticated VGG-like architecture with a slightly higher capacity than the MAE model, incorporating advanced techniques such as batch normalization, dropout, and a cosine learning rate scheduler. This strong baseline was explicitly optimized for high performance on the downstream task, providing a rigorous benchmark against which the effectiveness of the MAE method could be measured.

\begin{table}[ht]
\centering
\caption{Performance comparison of MAE models pre-trained on synthetic data (syn) and combined data (syn + real) against CNN baselines across varying downstream dataset sizes.}
\begin{tabularx}{\columnwidth} {c Y Y Y Y}
    \toprule
    \multicolumn{1}{c}{\textbf{Dataset}}
    & \multicolumn{2}{c}{\textbf{CNN (Baselines)}}
    & \multicolumn{2}{c}{\textbf{M-base Model}}\\    
    size & naive & strong & syn & syn + real \\
    \midrule
    12K 
    & 82.57 {\tiny$\!\pm$ 1.40}
    & 85.87 {\tiny$\!\pm$ 0.71}
    & \textbf{87.58} {\tiny$\!\pm$ 0.47}
    & 87.46 {\tiny$\!\pm$ 0.30}\\
    10K
    & 79.37 {\tiny$\!\pm$ 2.84}
    & 83.81 {\tiny$\!\pm$ 0.99}
    & \textbf{86.86} {\tiny$\!\pm$ 0.63}
    & 86.44 {\tiny$\!\pm$ 0.54}\\
    8K
    & 74.26 {\tiny$\!\pm$ 5.59}
    & 82.03 {\tiny$\!\pm$ 0.83}
    & 84.53 {\tiny$\!\pm$ 0.77}
    & \textbf{84.88} {\tiny$\!\pm$ 0.80}\\
    6K
    & 62.72 {\tiny$\!\pm$ 17.23}
    & 74.45 {\tiny$\!\pm$ 1.67}
    & \textbf{80.82} {\tiny$\!\pm$ 0.74}
    & 80.72 {\tiny$\!\pm$ 1.03}\\
    4K
    & 22.95 {\tiny$\!\pm$ 15.53}
    & 57.65 {\tiny$\!\pm$ 3.29}
    & 67.50 {\tiny$\!\pm$ 1.95}
    & \textbf{68.55} {\tiny$\!\pm$ 1.71}\\
    2K
    & 0.50 {\tiny$\!\pm$ 0.58}
    & 38.50 {\tiny$\!\pm$ 2.00}
    & 48.00 {\tiny$\!\pm$ 3.02}
    & \textbf{51.25} {\tiny$\!\pm$ 1.27}\\
   \bottomrule
\end{tabularx}
\label{tab:real-results}
\end{table}

The results clearly indicate that our MAE-based approach outperforms both the naive and strong baselines. With the full dataset of 12K~samples, our pre-trained MAE models—trained either solely on synthetic data or on a combination of synthetic and measured data achieve consistently higher accuracy than the strong CNN baseline (87.46\% and 87.58\% vs. 85.87\%). Although this margin is moderate at full data scale, the performance gap becomes more pronounced as the downstream dataset size decreases.

\Figure[t!](topskip=0pt, botskip=0pt, midskip=0pt)[width=0.999\columnwidth]{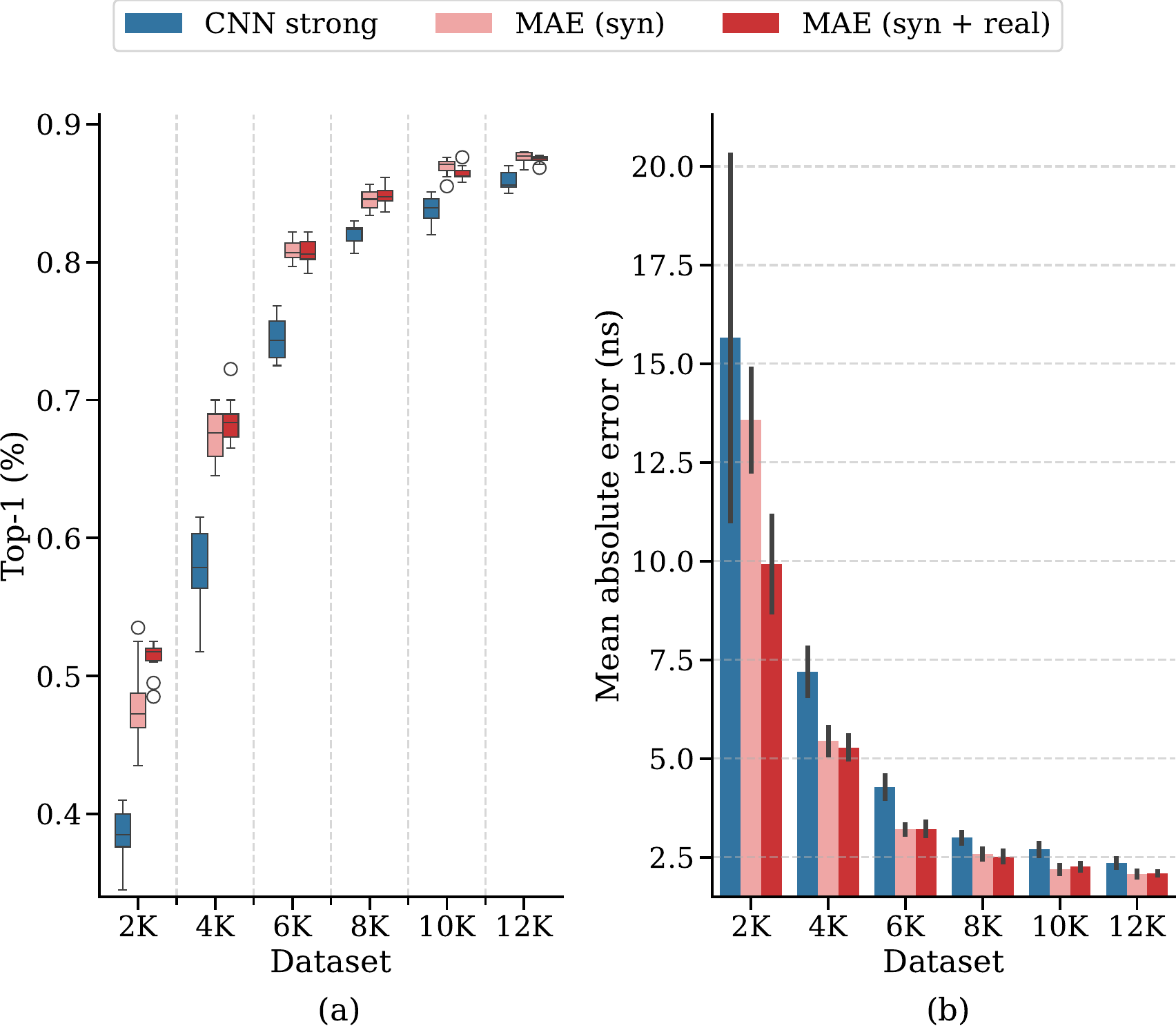}
{Comparison of MAE and strong CNN baseline across dataset sizes: (a) \mbox{top-1} accuracy and (b) mean absolute error.\label{fig:real-dataset-size}}

Notably, with only 6K~samples (half of the dataset), our MAE model pre-trained on synthetic data maintains high accuracy (80.82\%), significantly exceeding the strong CNN baseline (74.45\%). The performance gap increases further with smaller datasets. At 2K~samples, our MAE model achieves 48.00\% accuracy, and the combined pre-training model performs even better at 51.25\%, markedly surpassing the strong baseline at only 38.50\%. The naive CNN baseline struggles significantly, dropping to effectively equivalent to random guessing at this minimal dataset size.

These results highlight an important interplay between architectural inductive bias and the benefits of pre-training. While the strong CNN baseline likely benefits from its inherent inductive biases (e.g., spatial locality and translation equivariance), which are advantageous for signal processing tasks, especially when trained from scratch on limited labeled data, the pre-trained MAE model demonstrates the power of self-supervised learning. Despite the transformer's lower inductive bias, pre-training on a large unlabeled dataset allows it to learn highly effective representations directly from the data, ultimately surpassing the specialized CNN architecture. The significantly weaker performance of the MAE models trained from scratch further supports this interpretation, indicating that pre-training is crucial for leveraging the representational capacity of transformers on this task when labeled downstream data are scarce

These findings confirm that MAE pre-training yields robust representations, particularly under limited data conditions, and thus represents a promising approach for real-world ultrasound applications, particularly in scenarios with limited labeled data.

\subsection{Implementation Details}\label{sec:implementation-details}
For the training of all the models, a standardized setup was applied to ensure consistent and comparable results. A random masking strategy was used uniformly across all models, and learnable positional embeddings were employed instead of fixed sine-cosine functions, as both approaches have been shown to be beneficial for most applications. Each model was trained for $200$ epochs because further training did not yield significant performance improvements. The batch size was fixed at $1024$ because smaller batch sizes led to unstable training curves. Optimization was performed using an AdamW optimizer with a weight decay of $1 \times 10^{-4}$, $\beta_1 = 0.9$, and $\beta_2 = 0.999$.

Regularization techniques included a dropout rate of $0.1$, which was applied consistently in both the encoder and the decoder. Pre-Layer Normalization  was applied separately for Multi-Head Self-Attention and Multilayer Perceptron (MLP) layers within the encoder and decoder blocks. No data augmentation techniques were employed during training.

The training process for both the pre-training and downstream tasks utilized a cosine learning rate scheduler. For pre-training, the base learning rate was set to $1 \times 10^{-3}$, with a warm-up phase spanning $15\%$ of the total training steps. For the downstream tasks, the base learning rate was reduced to $0.05$, and the warm-up phase was shortened to $10\%$ of the total number of training steps. In both cases, the learning rate followed a cosine decay schedule after the warm-up phase and progressively decreased until the end of training.

The MLP within the encoder and decoder blocks was configured with two layers. The first layer projected the input to a dimensionality of $2 \times d_\text{model}$ and applied the Gaussian Error Linear Unit (GELU) activation function. The second layer reduces the dimensionality back to $d_\text{model}$ without an activation function.

Each model was trained ten times to account for instabilities observed during the training process, such as fluctuating loss curves or inconsistent convergence in certain configurations. The accuracy values and standard deviations reported in the tables reflect the mean performance and variability across the runs. Training was performed on a single NVIDIA A100 GPU.

\section{Conclusion}
This study investigates the application of Masked Autoencoders (MAE) with Vision Transformer (ViT) architectures to one-dimensional ultrasound signals, a challenging domain characterized by limited labeled data and the need for task-specific preprocessing. By leveraging pre-training on synthetic data, the approach achieves significant improvements in downstream tasks, such as ToF classification, compared with models trained from scratch.

Key findings highlight the impact of model size, patch size, masking ratio, and attention-head configuration on the overall performance. A masking ratio of 75\% combined with a patch size of $P = 32$ yielded the most effective results. Additionally, employing non-square query, key, and value matrices in the attention mechanism prevents overly narrow attention subspaces and further improves representation quality.

Notably, pre-training on synthetic data enhanced generalization to real-world measured signals, underlining the value of diverse and variable training data. This finding is further supported by comparative evaluations of both the naive and optimized CNN baselines. Although the performance margin over the strong CNN model is moderate when using the full dataset, the benefits of MAE become increasingly pronounced as the amount of labeled data decreases. This demonstrates the robustness and data efficiency of MAE representations, making the approach particularly suitable for practical ultrasound applications in which labeled data are limited, expensive, or difficult to obtain.

Notably, our findings suggest an advantage for pre-trained transformers over strong CNN baselines, even though CNNs possess inductive biases that are generally considered beneficial for signal data. This indicates that sufficient self-supervised pre-training enables transformer-based models such as MAE to overcome their lower inherent inductive bias by learning powerful, data-driven representations, ultimately achieving superior performance on downstream tasks such as ToF classification, especially in low-data regimes where training from scratch favors the biased CNNs.

Although the results of this study are promising, several limitations of the MAE approach should be acknowledged. First, although the reconstruction-based pre-training objective enables the model to learn broad and task-agnostic representations, unlike supervised learning, which tends to discard task-irrelevant information, its effectiveness across a diverse set of downstream tasks remains to be validated systematically. 

Furthermore, the MAE architecture introduces additional complexity compared to conventional CNNs, particularly when adapted to 1D ultrasound signals, where prior experience is limited, and standard implementations are optimized for 2D vision applications. In addition to the implementation effort, achieving strong performance requires extensive and systematic tuning of architecture-specific hyperparameters, most notably the masking ratio and patch size, which have a critical impact on representation quality and are not easily transferable across data modalities.

Overall, the findings show that MAE can be effectively adapted to 1D ultrasound data and offers a compelling strategy to reduce the dependence on large-scale labeled datasets.  Simultaneously, it helps reduce the need for manual feature engineering and task-specific signal processing by enabling general-purpose representation learning directly from raw signals.

Future work will focus on evaluating the transferability of learned representations across a broader range of downstream tasks, such as defect detection and material classification, to better understand the generalization capabilities of the MAE approach in real-world scenarios. A central objective is the drastic scaling of both the quantity and diversity of data used for pre-training. This includes systematically combining large volumes of synthetic signals with measured ultrasound data that exhibit high variability. To reflect realistic application conditions, the measured signals should be acquired under diverse settings, including variations in transducer types, propagation paths, and other boundary conditions. This strategy aims to expose the model to a broad spectrum of ultrasound time signals encountered in industrial environments with the goal of enabling the development of a robust foundation model for ultrasound signal analysis.

\section*{Acknowledgment}
This project was supported by Technologieallianz Oberfranken (TAO).

\bibliographystyle{IEEEtran} 
\bibliography{references} 

\begin{IEEEbiography}[{\includegraphics[width=1in,height=1.25in,clip,keepaspectratio]{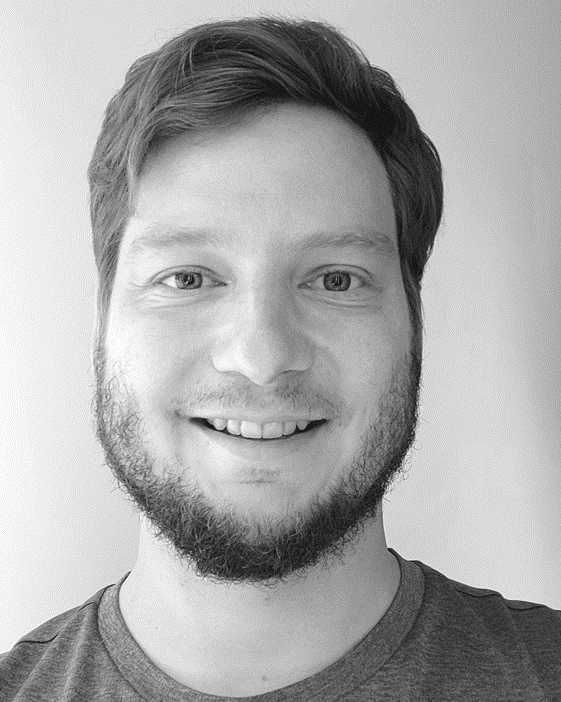}}]{Immanuel Roßteutscher}
received the Dipl.-Ing. (FH) degree in physical engineering in 2012 and the M.Eng. degree in electrical engineering and information technology in 2015, both from the Coburg University of Applied Sciences and Arts, Coburg, Germany.

He is currently a Research Associate with the Institute for Sensor and Actuator Technology (ISAT), Coburg University of Applied Sciences and Arts, where he is pursuing his Ph.D. degree. His research interests include ultrasound applications in industrial environments, digital signal processing, and the application of deep learning techniques for the analysis of ultrasound signals.
\end{IEEEbiography}

\begin{IEEEbiography}[{\includegraphics[width=1in,height=1.25in,clip,keepaspectratio]{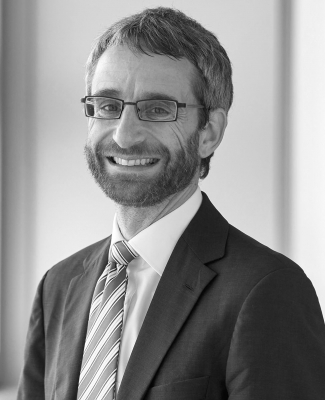}}]{Klaus S. Drese}
received the Diplom Physiker (Diploma in Physics) degree from the University of Würzburg, Germany, in 1995, and the Master of Arts degree from the State University of New York at Stony Brook, USA, in 1993. He received his Ph.D. degree in statistical physics from the University of Marburg, Germany, in 1998, under the supervision of Prof. Dr. Sigfried Großmann, with a dissertation on the control of simple quantum mechanical systems by laser radiation.

Following his doctorate, he held scientific positions at the University of Marburg and at IMM GmbH, where he later became Head of the Simulation Group and Head of the Department Fluidic and Simulation. He subsequently served as Scientific Director for micro analytical systems at IMM GmbH and later as Scientific Director and Head of Department Future Technologies at Fraunhofer ICT-IMM. Since October 2016, he has been a Professor for Sensorics, Microfluidics, and Microacoustics in the Faculty of Applied Natural Sciences and Health at the Coburg University of Applied Sciences and Arts, Germany. He also serves as the Director of the Institute of Sensor and Actuator Technology (ISAT) at Coburg University. His research interests include microfluidics, sensor technology, simulation techniques, and microacoustics. 

Prof.\ Drese is an active member of the research community, serving on editorial boards for journals such as "Sensors" and "Micromachines," and is involved in the organization of scientific conferences. He is a member of the German Physical Society and the AMA Association for Sensor and Measurement.
\end{IEEEbiography}

\begin{IEEEbiography}[{\includegraphics[width=1in,height=1.25in,clip,keepaspectratio]{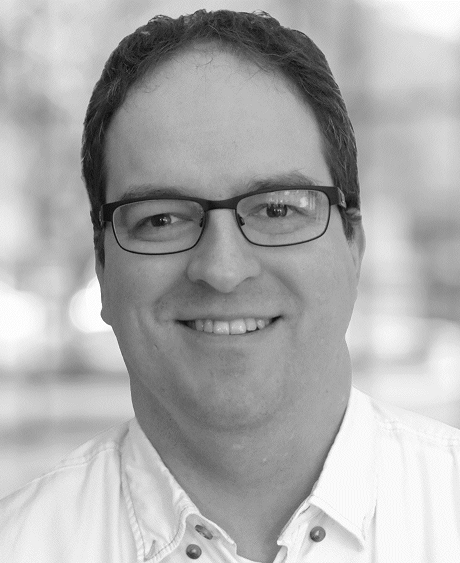}}]{Thorsten Uphues} received the Diploma and Ph.D. degrees in physics from the University of Bielefeld, Germany, in 2003 and 2007, respectively, under the supervision of Prof. Dr. DrSc. h.c. Ulrich Heinzmann. His doctoral work focused on attosecond-time-resolved ionization chronoscopy. 

He conducted postdoctoral research at the Max-Planck-Institute of Quantum Optics in the department of the Nobel laureate in Physics 2023, Prof. Ferenc Krausz, contributing to foundational work in attosecond science. From 2008 to 2012, he worked in applied analytics and project management at empolis GmbH and Attensity Europe GmbH. From 2012 to 2018, he served as Junior Professor of Experimental Physics at the University of Hamburg and the Center for Free-Electron Laser Science CFEL. Since 2020, he has been a Professor in Physics and Director of the Institute for Sensor and Actuator Technology (ISAT), Coburg University of Applied Sciences and Arts, Germany. His research interests include ultrafast laser physics, industrial sensor technology, and the integration of artificial intelligence into physical measurement systems. He has authored numerous publications in leading journals in applied physics and photonics, and is actively involved in collaborative, interdisciplinary research projects. 

Prof.\ Uphues is a Senior Member of OPTICA, a Full Member of the Sigma Xi Society, and a Member of SPIE. He received the Otto-Hahn Medal of the Max Planck Society for his contributions to time-resolved X-ray physics.
\end{IEEEbiography}

\EOD
\end{document}